\documentclass{article}
\usepackage{spconf,amsmath,graphicx}
\usepackage{enumitem}
\usepackage{amsmath, color, soul}
\usepackage{amssymb}
\usepackage{multirow}
\usepackage{pifont}
\usepackage{xcolor}
\usepackage{hyperref}

\definecolor{Agriculture}{RGB}{154, 205, 50} 
\definecolor{Rangeland}{RGB}{0, 255, 255}   
\definecolor{Forest}{RGB}{0, 255, 0}        
\definecolor{Water}{RGB}{0, 0, 255}         
\definecolor{Barren}{RGB}{255, 0, 0}        
\definecolor{Urban}{RGB}{255, 0, 255}       

\newcommand{\cmark}{\ding{51}}%
\newcommand{\xmark}{\ding{55}}%

\title{Source-Free Online Domain Adaptive Semantic Segmentation of Satellite Images under Image Degradation}
\name{Fahim Faisal Niloy$^{\dagger}$ \qquad Kishor Kumar Bhaumik$^{\ddagger}$ \qquad Simon S. Woo$^{\ddagger}$ \thanks{Project page: \href{https://sat-tta.github.io}{https://sat-tta.github.io}}}
  
\address{$^{\dagger}$University of California, Riverside \\ $^{\ddagger}$Sungkyunkwan University, South Korea}
\begin{document}
\maketitle
\begin{abstract}
Online adaptation to distribution shifts in satellite image segmentation stands as a crucial yet underexplored problem. In this paper, we address source-free and online domain adaptation, i.e., test-time adaptation (TTA), for satellite images, with the focus on mitigating distribution shifts caused by various forms of image degradation. Towards achieving this goal, we propose a novel TTA approach involving two effective strategies. First, we progressively estimate the global Batch Normalization (BN) statistics of the target distribution with incoming data stream. Leveraging these statistics during inference has the ability to effectively reduce domain gap. Furthermore, we enhance prediction quality by refining the predicted masks using global class centers. Both strategies employ dynamic momentum for fast and stable convergence. Notably, our method is backpropagation-free and hence fast and lightweight, making it highly suitable for on-the-fly adaptation to new domain. Through comprehensive experiments across various domain adaptation scenarios, we demonstrate the robust performance of our method.
\end{abstract}
\begin{keywords}
Satellite Image, Test-Time Adaptation, Domain Adaptation, Segmentation
\end{keywords}
\section{Introduction}


In real-world satellite image segmentation, numerous situations can arise that require domain adaptation. For instance, variations between satellite images taken from different cities or diverse sensor noise characteristics from different satellites can lead to discernible domain disparities. These disparities can hinder a model trained on a specific dataset from generalizing effectively to target datasets. Unsupervised domain adaptation (UDA) \cite{chen2023s, Tasar_2020_CVPR_Workshops} helps bridge these gaps but typically requires both source and target data, which isn't always feasible due to privacy and storage constraints. To address this, methods like \cite{xu2023universal, xu2022source} have been proposed, allowing adaptation without the need for source data, known as source-free domain adaptation (SFDA).

Both UDA and SFDA require the entire target domain data to be available during adaptation. Additionally, the adaptation is performed offline, meaning that it occurs before the model is deployed. However, many situations might arise when online adaptation is necessary, especially in cases of domain shifts during the online segmentation of satellite images using the deployed model. For example, during natural disasters such as earthquakes, floods, and wildfires, online segmentation is crucial for quickly identifying affected areas, assessing damage, and prioritizing rescue and relief efforts. Online segmentation can also be used to monitor crop health, detect disease outbreaks, and estimate yields. However, such online segmentation scenarios can face major challenges due to the degradation of satellite images caused by various factors, including different characteristics of sensor noise, atmospheric and weather conditions (e.g., rain, fog). Hence, the development of a model that can adapt to such domain shifts on-the-fly and without having access to source data becomes crucial. Such source-free and online adaptation is also termed as test-time adaptation (TTA) in the literature \cite{wang2020tent}.



In this paper, we propose a novel TTA approach for satellite image segmentation. Previous research \cite{li2016revisiting} has shown that utilizing Batch Normalization (BN) statistics during inference instead of the source pre-trained statistics can effectively mitigate domain gap. By employing a running average method with dynamic momentum, we accurately approximate target BN statistics. This significantly reduces the domain gap and enhances performance compared to source models. Furthermore, we take a running average of confident pixels to keep track of global class centers to further refine the prediction map. Importantly, our method does not require any backpropagation during adaptation; hence, it is fast and lightweight. This feature is highly desirable in scenarios like TTA of satellite images, where images are typically high-resolution and can take up a significant amount of storage. A lightweight adaptation method ensures it does not impose additional memory constraints.

Moreover, an additional limitation in the current domain adaptation research for satellite images is that it mainly focuses on adapting between different geographic regions, overlooking domain gaps that arise from various sources of degradation such as sensor noise, weather conditions, etc. Addressing such domain gaps is even more crucial during online adaptation, as mentioned earlier. The primary reason behind this is the scarcity of datasets dedicated to these specific forms of degradation. To address the gap, we introduce a new synthetic satellite image dataset by simulating different forms of image degradation. In summary, our contributions are:


\noindent
\begin{itemize}[leftmargin=*,topsep=0pt]
\setlength{\itemsep}{-2pt}
\item We address a novel problem of source-free and online domain adaptive semantic segmentation of satellite images under various forms of image degradation. To the best of our knowledge, this is the first work to explore such setting for satellite image segmentation.
\item We present a novel TTA method that involves approximating the target BN statistics and global class centers with incoming data stream using dynamic momentum value.
\item We evaluate our method on the TTA of satellite images and demonstrate that our method outperforms state-of-the-art generic TTA approaches on this task.
\end{itemize}




\section{Background}

\textbf{Related Works: }  Previous research \cite{chen2023s, Tasar_2020_CVPR_Workshops, 9070187} have examined UDA for satellite images. Notable methods include generative approaches \cite{Tasar_2020_CVPR_Workshops}, discrepancy-based approaches \cite{9070187}, and self-training \cite{10130291}. UDA for satellite images has also been extended to the source-free case by \cite{xu2023universal, xu2022source}. 
Recently, many generic TTA methods have been proposed for both classification and segmentation tasks. In particular, TENT \cite{wang2020tent} adapts a pre-trained source model on incoming target data by minimizing entropy and updating the Batch-Norm parameters of the source model. On the other hand, DUA \cite{mirza2022norm} continuously updates the BN statistics of the pre-trained source model with the incoming test batches in order to align to the target distribution. Also, DIGA \cite{wang2023dynamically} mixes source statistics with incoming target statistics for adaptation. 




\noindent \textbf{Problem Definition:} Consider a segmentation model $f_{{\theta}_{0}}$ pre-trained on the source image data $\mathcal{X}_{s} \sim \mathcal{D}_s$, where $\mathcal{D}_s$ denotes the source distribution. During deployment, the model encounters a sequence of test data $\mathcal{X}_{1} \rightarrow \mathcal{X}_{2}\rightarrow \ldots \rightarrow \mathcal{X}_{t} \rightarrow \ldots$, where $\mathcal{X}_{t}$ represents a small batch of test samples from the source degraded stationary target distribution $\mathcal{D}_{\textit{T}}$. Following the TTA setting, the model only has access to $\mathcal{X}_{t}$ at an instance. Furthermore, the model needs to adapt to each incoming test batch $\mathcal{X}_{t}$ and update its parameters from $f_{{\theta}_{t-1}} \rightarrow f_{{\theta}_{t}}$ in order to improve its output on the subsequent test batches. 

\section{Method}
\begin{table*}[t!]
\centering
\small
\caption{Adaptation Performance in IoU (\%) of our method and baseline methods. The model is pre-trained on the clean and uncorrupted training set and during test-time, adapted to the source-degraded target domain in an online manner.}
\label{tab:main_table}
\begin{tabular}{ccccccccc}
\hline
\multirow{2}{*}{Task} &
  \multirow{2}{*}{Method} &
  \multicolumn{6}{c}{IoU (\%)} &
  \multirow{2}{*}{mIoU (\%)} \\ \cline{3-8}
 &
   &
  Agriculture &
  Rangeland &
  Forest &
  Water &
  Barren &
  Urban &
   \\ \hline
\multirow{5}{*}{\begin{tabular}[c]{@{}c@{}}Clean \\ $\downarrow$ \\ Gaussian\\ Noise \end{tabular}} &
  Source-Only 
& 32.4
& 14.3
& 37.5
& 42.2
& 29.0
& 55.8
& 35.2
   \\
 &
  TENT \cite{wang2020tent}
  & 74.5
& 16.9
& 60.8
& 29.8
& 46.8
& 60.1
& 48.2
   \\
 &
  DUA \cite{mirza2022norm}
& 67.3
& 19.7
& 52.8
& 47.8
& 48.4
& 63.9
& 50.0
   \\
 &
  DIGA \cite{wang2023dynamically} 
  & 75.8
& 16.6
& 62.2
& 29.4
& 54.1
& 62.4
& 50.1
   \\
 &
  \textbf{OURS}
  & 77.3
& 18.6
& 66.5
& 45.0
& 52.2
& 64.0
& \textbf{53.9}
   \\ \hline
\multirow{5}{*}{\begin{tabular}[c]{@{}c@{}}Clean\\ $\downarrow$\\ Impulse\\ Noise\end{tabular}} &
  Source-Only 
  & 37.2
& 15.3
& 47.0
& 40.7
& 27.3
& 56.4
& 37.3
   \\
 &
  TENT 
  & 75.3
& 18.0
& 64.5
& 31.3
& 48.7
& 61.5
& 49.9
   \\
 &
  DUA 
  & 71.1
& 20.1
& 61.1
& 44.2
& 47.8
& 64.9
& 51.5
   \\
 &
  DIGA 
  & 75.2
& 20.0
& 61.9
& 25.1
& 52.7
& 63.3
& 49.7
   \\
 &
  \textbf{OURS} 
  & 78.1
& 22.9
& 67.8
& 48.6
& 54.0
& 65.5
& \textbf{56.1}
   \\ \hline
\multirow{5}{*}{\begin{tabular}[c]{@{}c@{}}Clean\\ $\downarrow$\\ Gaussain\\ Blur\end{tabular}} &
  Source-Only 
  & 78.3
& 16.8
& 66.2
& 54.9
& 57.9
& 62.2
& 56.1
   \\
 &
  TENT 
  & 78.6
& 19.5
& 68.4
& 61.6
& 58.2
& 60.9
& 57.9
   \\
 &
  DUA 
  & 80.3
& 22.7
& 68.5
& 58.5
& 60.1
& 66.1
& 59.4
   \\
 &
  DIGA
  & 80.5
& 16.1
& 70.9
& 64.6
& 59.5
& 64.1
& 59.3
   \\
 &
  \textbf{OURS} 
  & 81.4
& 23.9
& 73.3
& 67.5
& 61.3
& 64.6
& \textbf{62.0}
   \\ \hline
\multirow{5}{*}{\begin{tabular}[c]{@{}c@{}}Clean\\ $\downarrow$\\ Snow\end{tabular}} &
  Source-Only 
  & 0.0
& 0.0
& 6.5
& 3.7
& 2.1
& 8.7
& 3.5

   \\
 &
  TENT 
  & 68.5
& 14.8
& 59.0
& 20.3
& 32.4
& 46.2
& 40.2 \\
 &
  DUA 
  & 11.0
& 9.0
& 34.9
& 10.2
& 15.2
& 22.2
& 17.1
  \\
 &
  DIGA 
  & 50.5
& 12.6
& 57.0
& 14.7
& 24.2
& 48.3
& 34.5
   \\
 &
  \textbf{OURS} 
  & 72.1
& 16.2
& 62.1
& 32.7
& 42.1
& 49.1
& \textbf{45.7}
   \\ \hline
\multirow{5}{*}{\begin{tabular}[c]{@{}c@{}}Clean\\ $\downarrow$\\ Fog\end{tabular}} &
  Source-Only 
  & 11.7
& 6.4
& 8.9
& 4.4
& 18.3
& 25.6
& 12.6
   \\
 &
  TENT 
  & 72.2
& 15.4
& 45.9
& 31.6
& 37.3
& 47.1
& 41.6
   \\
 &
  DUA
  & 51.0
& 15.3
& 32.3
& 10.3
& 31.4
& 39.6
& 30.0
  \\
 &
  DIGA 
  & 68.0
& 14.4
& 45.4
& 38.8
& 37.5
& 45.2
& 41.6
   \\
 &
  \textbf{OURS} 
  & 75.6
& 18.6
& 57.4
& 43.9
& 43.9
& 51.9
& \textbf{48.6}
   \\ \hline
\end{tabular}
\end{table*}
Our main objective is to achieve efficient online adaptation without imposing significant additional resources. To meet this objective, we employ two backpropagation-free strategies that significantly enhance performance in the new domain.

\noindent \textbf{Distribution Matching:} Image degradation can cause the feature distribution of target domain to be different from source distribution. Hence, we need to align both the distribution for adaptation. Li et al. \cite{li2016revisiting} have demonstrated that utilizing the BN statistics of the target domain during inference instead of the pre-trained source statistics can effectively mitigate domain gap. However, TTA usually assumes very small batch size per instance for adaptation. Hence, approximating the correct BN statistics for target distribution with such small amount of data per instance
is difficult. To address the problem, we utilize running average of BN statistics to progressively approximate the global BN statistics of the incoming domain. Specifically, during testing on the current batch $t$, let $\tilde{\mu}_{l_{c}}$ and $\tilde{\sigma}_{l_{c}}$ represent the BN statistics (mean and variance) computed from the $l$-th layer's $c$-th channel of the model. We calculate two running averages as follows:
\begin{align}
    \mu_{l_{c}}^{t} &= (1-\alpha^{t})\times\mu_{l_{c}}^{t-1} + \alpha^{t}\times \tilde{\mu}_{l_{c}} \label{eq:running_mu}\\
    \sigma_{l_{c}}^{t} &= (1-\alpha^{t})\times\sigma_{l_{c}}^{t-1} + \alpha^{t}\times \tilde{\sigma}_{l_{c}}^{t} \ . \label{eq:running_sigma}
\end{align}
Here, $\mu_{l_{c}}^{t}$ and $\sigma_{l_{c}}^{t}$ denote the running BN statistics that get updated with each incoming test batch. We initialize $\mu_{l_{c}}^{0}$ and $\sigma_{l_{c}}^{0}$ with the BN statistics of the pre-trained source model, and $\alpha^{t}$ is the momentum value that controls the effect of current batch statistics on the running statistics.

In fact, designing the momentum term accurately is important. While a constant value may serve its purpose, it weighs all the incoming test batches equally to update the running BN statistics. A high value of momentum renders the running stats sensitive to outliers. On the other hand, setting a small constant value for the momentum, will result in the running stats to converge slowly. To overcome these challenges, we adopt a dynamic approach for the momentum, gradually reducing its value after each iteration according to the formula:
\begin{align}
    \alpha^{t} = \alpha^{t-1} \times \gamma_{dm}
    \label{eq:running_sigma}
\end{align}
Here, $\gamma_{dm}$ is a constant in the interval $(0, 1)$. This dynamically decreasing momentum ensures that the running statistics eventually stabilize to a constant value, serving as an approximation of the global BN statistics for the target domain.




\noindent \textbf{Instance Matching:} Corruption or degradation can cause a notable divergence in the visual attributes of specific individual pixels in target images compared to the source image, potentially leading to inaccurate predictions. To address the problem,  inspired by previous works \cite{wang2023dynamically, snell2017prototypical, pan2019transferrable} that use prototype-based methods, we calculate a running average of the features of confident classes and leverage them to guide the recognition of wrongly recognized pixels. However, unlike previous approaches, we use a dynamic momentum value that helps for faster and more stable convergence. 

Specifically, let the output logits of the segmentation model after correcting for BN Statistics, be denoted as $F \in \mathbb{R}^{C \times H \times W}$, where $C$, $H$, and $W$ denote the channels, height, and width of the feature map, respectively. The classifier takes $F$ to predict the softmax output $L \in \mathbb{R}^{K \times H \times W}$, where $K$ is the number of classes. The straightforward definition of a region or class center of class $i$ for the current sample can be defined as:
\begin{align}
     \tilde{R}_{i} = \frac{\sum_{x,y}F_{(x,y)} \mathbb{I} [ \text{argmax}(L_{(x,y)}) = i, \text{max}(L_{(x,y)}) \geq \mathcal{P}_{0}] } {\sum_{x,y} \mathbb{I} [\text{argmax}(L_{(x,y)}) = i, \text{max}(L_{(x,y)}) \geq \mathcal{P}_{0}]}
\end{align}
where $\mathbb{I}(\cdot)$ is the binary indicator denoting whether the pixel belongs to class $i$ and the predicted probability for that class is greater than $\mathcal{P}_{0}$. For each class center, we compute a running average to capture the global class center prototype of that particular domain. Define $R_{i}^{t}$ as follows:
\begin{align}
    R_{i}^{t} &= (1-\beta^{t})\times R_{i}^{t-1} + \beta^{t}\times \tilde{R_{i}} 
\end{align}
Here also, we use a dynamic momentum value for $\beta$:
\begin{align}
    \beta^{t} = \beta^{t-1} \times \gamma_{im}
\end{align}
Given the class centers, we can calculate the instance-matched prediction $P \in \mathbb{R}^{K \times H \times W}$ as follows:
\begin{align}
   P_{(K=i,x,y)} = \frac{\text{exp} (-<F_{(x,y)}, R_{i}^{t}>)} {\sum_{K \neq i} \text{exp} (-<F_{(x,y)}, R_{i}^{t}>)}
\end{align}
Then, the final prediction mask $M$ can be computed by weighting with $\gamma$:
\begin{align}
   M_{(x,y)} = (1-\gamma) \times L_{(x,y)} + \gamma \times P_{(x,y)}
\end{align}
\section{Experiments}
\textbf{Dataset:}
We use the benchmark satellite dataset DeepGlobe \cite{demir2018deepglobe} as our primary dataset. To simulate some of the most common image corruptions observed in satellite images, such as Gaussian Noise, Impulse Noise (commonly known as Salt and Pepper Noise), Gaussian Blur, Fog, and Snow, we utilize the RobustBench \cite{croce2020robustbench} simulation. These corruptions are applied to the DeepGlobe test set. 

\noindent \textbf{Baseline Models:} For the baseline models, we have selected state-of-the-art generic TTA algorithms - TENT \cite{wang2020tent}, DUA \cite{mirza2022norm}, and DIGA \cite{wang2023dynamically}. The source codes for all the baseline methods are publicly available. We also report the Source-Only performance, which is the result obtained from the evaluation of the source model on the target distribution without performing adaptation.

\noindent \textbf{Implementation Details:} We have utilized DeepLab-v3 \cite{chen2017rethinking} with a ResNet-50 backbone as our segmentation model. The hyperparameters we have used include $\alpha^{0} = \beta^{0} = 0.9$, $\gamma_{dm}=\gamma_{im}=0.95$, $\gamma=0.2$, and $\mathcal{P}_{0}=0.5$. DeepGlobe images are resized to a shape of $512 \times 512$. Our batch size is set to $8$. As for the baseline methods, we have followed the hyperparameter choices specified in their respective papers.

\begin{figure}[t!]
\setlength\tabcolsep{1pt}
\centering
\begin{tabular}{ccccc}
\includegraphics[width=0.8in]{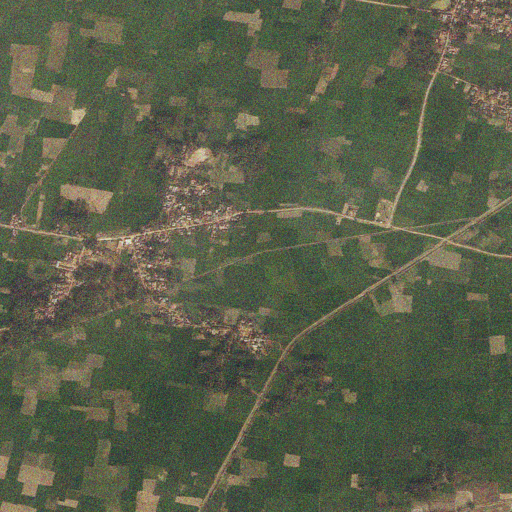} &
 \includegraphics[width=0.8in]{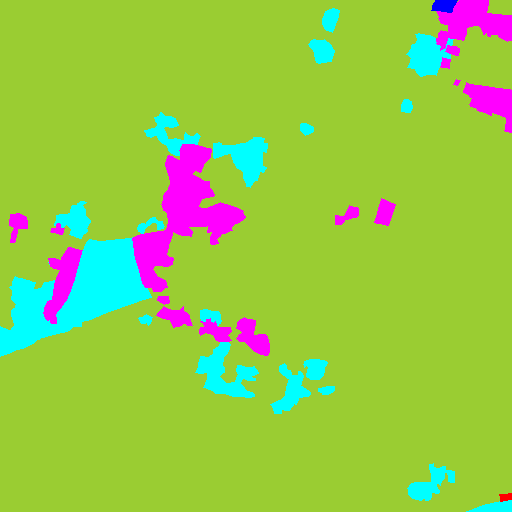} &
 \includegraphics[width=0.8in]{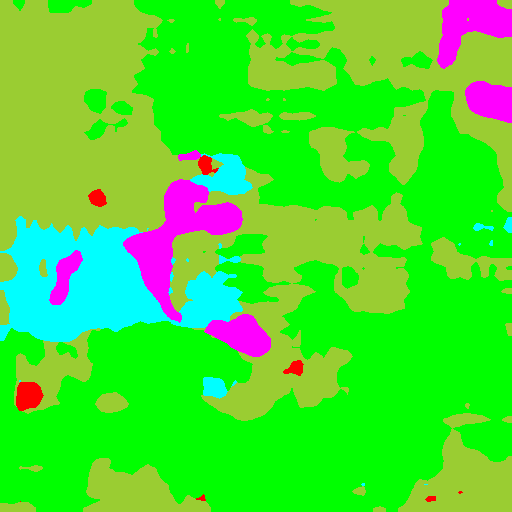} &
 \includegraphics[width=0.8in]{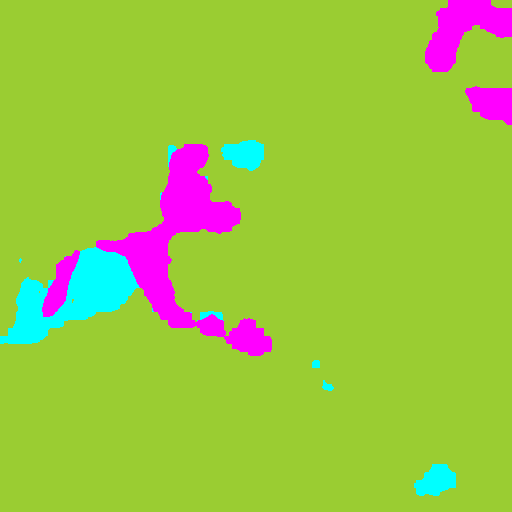} \\
 \includegraphics[width=0.8in]{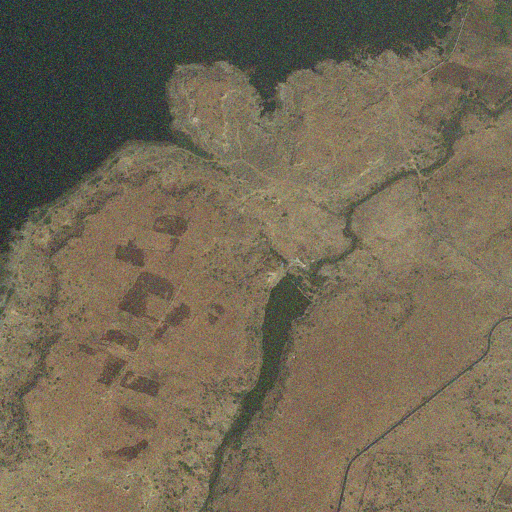} &
 \includegraphics[width=0.8in]{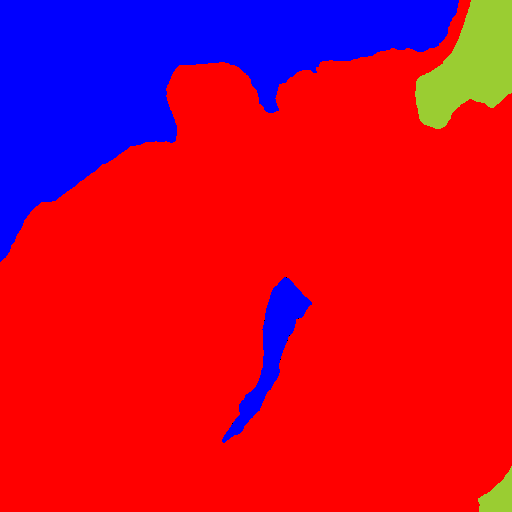} &
 \includegraphics[width=0.8in]{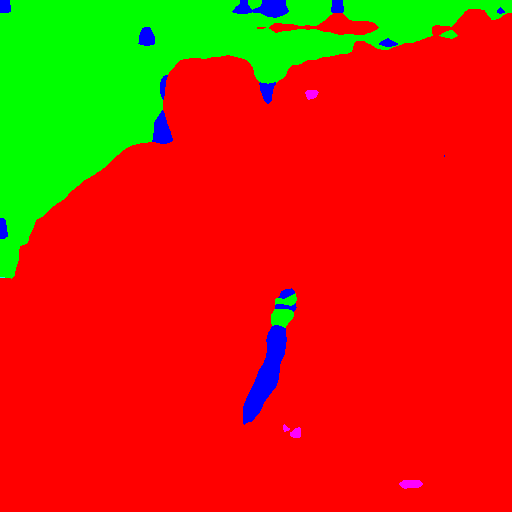} &
 \includegraphics[width=0.8in]{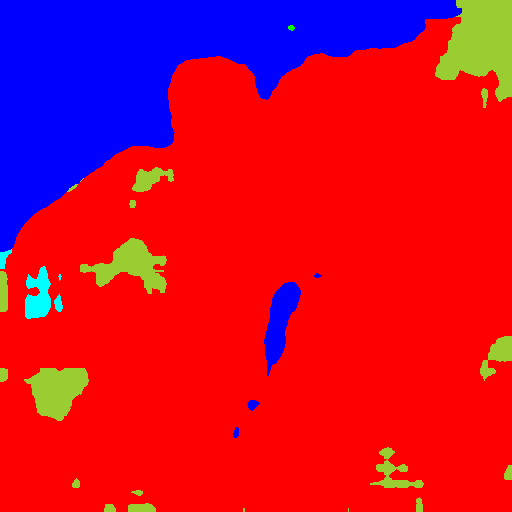} \\
 \includegraphics[width=0.8in]{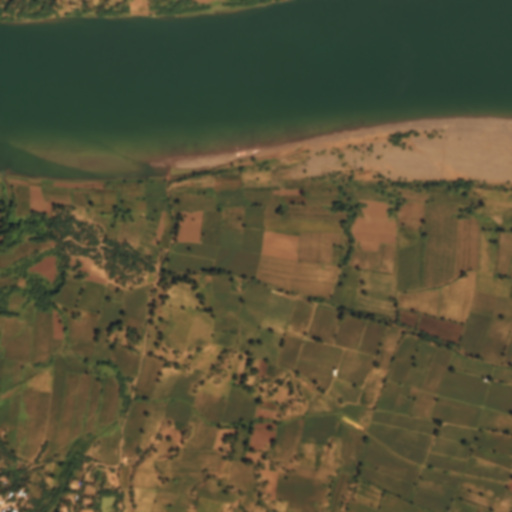} &
 \includegraphics[width=0.8in]{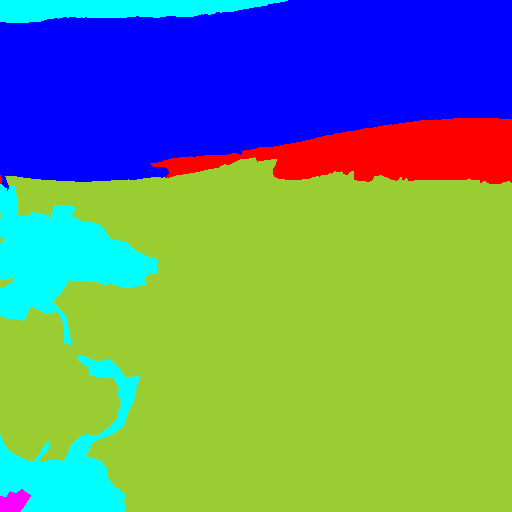} &
 \includegraphics[width=0.8in]{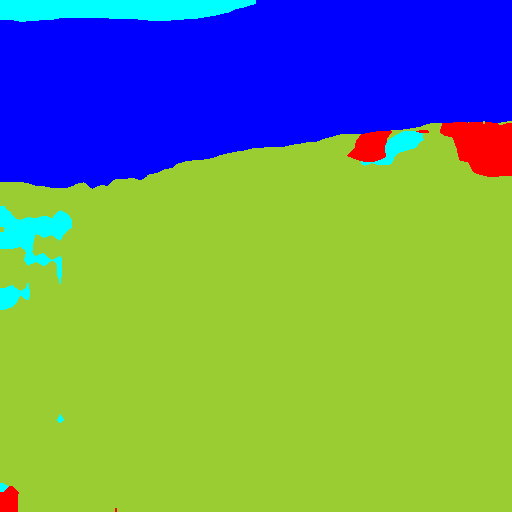} &
 \includegraphics[width=0.8in]{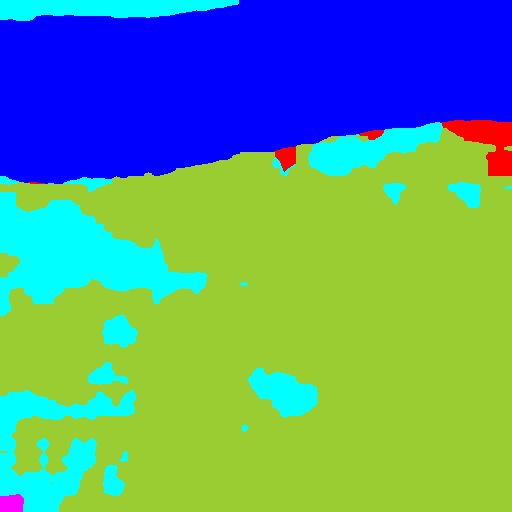}  \\
 \includegraphics[width=0.8in]{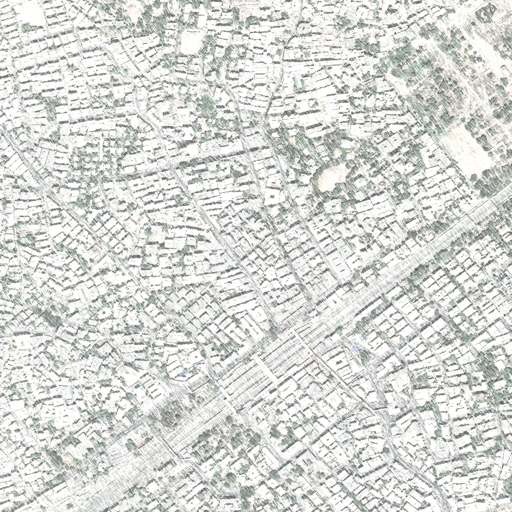} &
 \includegraphics[width=0.8in]{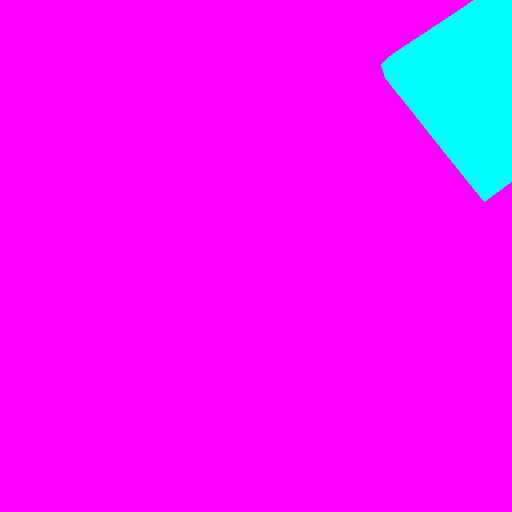} &
 \includegraphics[width=0.8in]{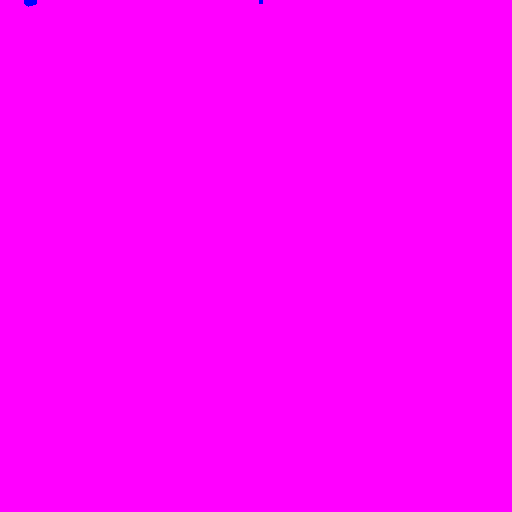} &
 \includegraphics[width=0.8in]{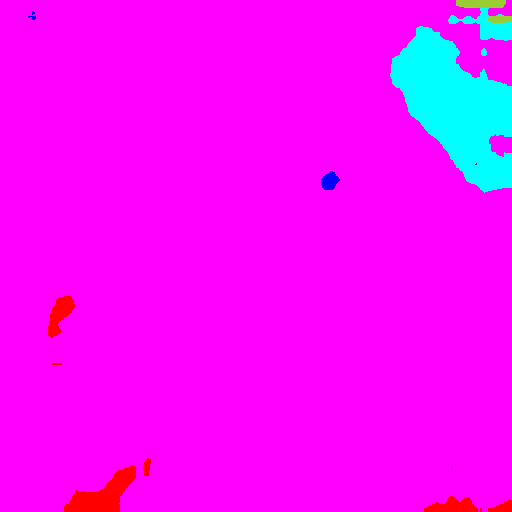} \\
\includegraphics[width=0.8in]{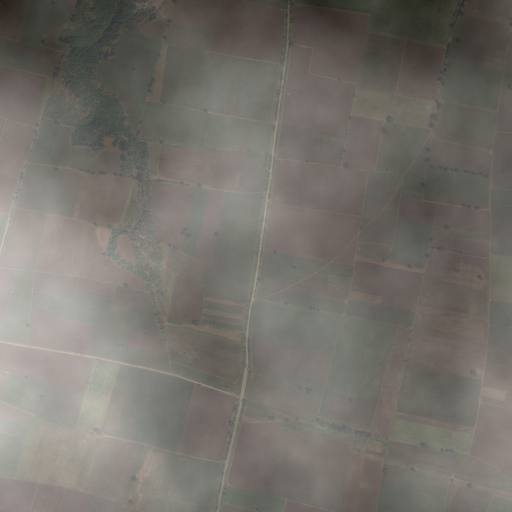} &
 \includegraphics[width=0.8in]{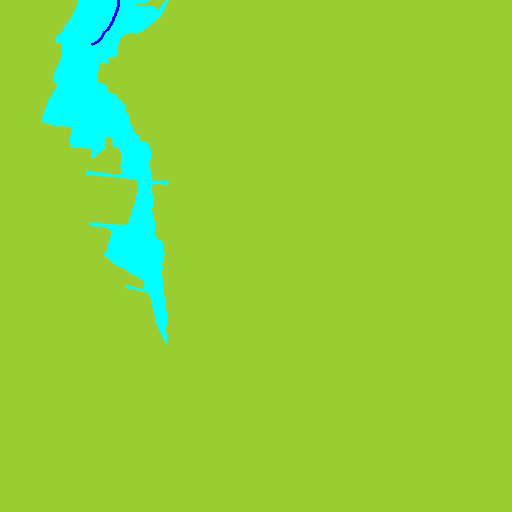} &
 \includegraphics[width=0.8in]{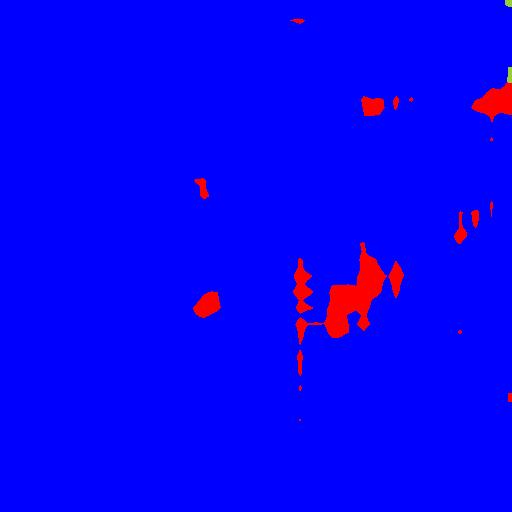} &
 \includegraphics[width=0.8in]{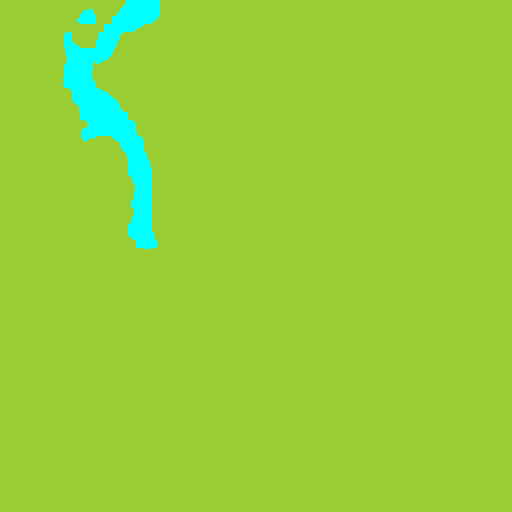} \\
 \textbf{\footnotesize{Input}} &
 \textbf{\footnotesize{GT Mask}} &
 \textbf{\footnotesize{Source-Only}} &
 \textbf{\footnotesize{Ours}} \\
\end{tabular}
\caption{Comparison of the predicted mask of our method with the Source-Only model. The rows correspond to examples of Gaussian Noise, Impulse Noise, Gaussian Blur, Snow and Fog from top to bottom respectively. Color-coded legends: \textcolor{Agriculture}{\rule[0.5ex]{1em}{2.5pt}}  (Agriculture), \textcolor{Rangeland}{\rule[0.5ex]{1em}{2.5pt}} (Rangeland),  \textcolor{Forest}{\rule[0.5ex]{1em}{2.5pt}} (Forest), \textcolor{Water}{\rule[0.5ex]{1em}{2.5pt}} (Water), \textcolor{Barren}{\rule[0.5ex]{1em}{2.5pt}} (Barren), \textcolor{Urban}{\rule[0.5ex]{1em}{2.5pt}} (Urban).}
\label{fig:segmentation}
\end{figure}

\noindent \textbf{Results:} Our adaptation performance is presented in Table \ref{tab:main_table}. Initially, the source model is trained on the clean and uncorrupted DeepGlobe train set. During testing, we sequentially adapt the clean model to each batch of target images from the target domain involving image degradation. As observed in Table \ref{tab:main_table}, degradation can severely impact the performance of the Source-Only model, even to the extent where the source performance drops to as low as $3.5\%$, particularly in the case of snow. This is a significant drop considering the fact that the source model achieves an mIoU of $62.9\%$ on the clean and uncorrupted DeepGlobe test set. This phenomenon underscores the critical need for adaptation in the context of degraded satellite images. Furthermore, Table \ref{tab:main_table} demonstrates that in all cases, our method has consistently outperformed all the baseline models in terms of mean performance. We also provide visualizations of the predicted segmentation masks generated by our method in Figure \ref{fig:segmentation}. For comparison purposes, we show the predictions of the Source-Only model. It is evident that the naive source-only prediction, without adaptation, performs poorly. Conversely, our method generates output masks that closely resemble the actual ground truth.

\noindent \textbf{Ablation Study:}
Here, we aim to demonstrate the effectiveness of the two strategies employed in our method. Specifically, we focus on the Clean to Impulse Noise adaptation task, and the results are presented in Table \ref{tab:ablation}. It clearly illustrates that both strategies are individually effective in adaptation. Furthermore, combining the results of both methods leads to even greater improvements in performance.

\begin{table}[h!]
\centering
\caption {Ablation study for the distribution matching (DM) and instance matching (IM) strategies used in our method.}
\label{tab:ablation}
\begin{tabular}{ccc}
\hline
DM & IM & mIoU (\%) \\ \hline
\xmark  & \xmark  & 37.3 \\
\xmark  & \cmark  & 50.7 \\
\cmark  & \xmark  & 54.8 \\
\cmark  & \cmark  & \textbf{56.1} \\ \hline
\end{tabular}
\end{table}
In our method, we have employed a dynamic momentum value for both $\alpha$ and $\beta$. We now demonstrate the effectiveness of using dynamic momentum over a fixed value for both parameters.

\begin{table}[h!]
\centering
\caption{Efficacy of dynamic momentum.}
\label{tab:ablation_dyn}
\begin{tabular}{ccccc}
\hline
Momentum & 0.1  & 0.5  & 0.9  & Ours          \\ \hline
mIoU (\%)     & 34.8 & 51.4 & 47.3 &  \textbf{56.1} \\
\hline
\end{tabular}
\end{table}

As shown in Table \ref{tab:ablation_dyn}, it is clear that using dynamic momentum is more effective than using a constant value. Additionally, constant values have the disadvantage of being sensitive to outliers and may require fine-tuning for different datasets based on domain distance. In contrast, dynamic momentum is robust to these disadvantages and ensures a faster and stable convergence.

\section{Conclusion}
In this paper, we have addressed the problem of source-free and online adaptation of satellite images to distribution shifts caused by various forms of image degradations. To address this issue, we have presented a novel TTA method that utilizes the running average of target BN statistics to approximate the global BN statistics. Additionally, we employ global class centers to further refine the segmentation masks. Our method is backpropagation-free and lightweight, making it well-suited for online adaptation of satellite images. Through various experiments, we have demonstrated that our approach outperforms several state-of-the-art generic TTA methods in this task.

\noindent \textbf{Acknowledgment.} This work was supported by Institute of Information \& communications Technology Planning \& Evaluation (IITP) grant by the Korea government (MSIT) (No.RS-2023-00230337,Advanced and Proactive AI Platform Research and Development Against Malicious Deepfakes).



\bibliographystyle{IEEEbib}
\bibliography{strings,refs}

\end{document}